\documentclass[letterpaper, 10 pt, journal, twoside]{IEEEtran}
\usepackage{cite}
\usepackage{amsmath,amssymb,amsfonts}
\usepackage{algorithmic}
\usepackage{graphicx}
\usepackage{textcomp}
\usepackage{xcolor}

\usepackage{booktabs}
\usepackage{geometry}
\usepackage{longtable}
\usepackage{caption}
\usepackage{subcaption}
\usepackage{flushend}
\usepackage[
  separate-uncertainty = true,
  multi-part-units = repeat
]{siunitx}

\usepackage{gensymb}

\makeatletter %
\newcommand{\linebreakand}{%
  \end{@IEEEauthorhalign}
  \hfill\mbox{}\par
  \mbox{}\hfill\begin{@IEEEauthorhalign}
}
\makeatother %

\def\BibTeX{{\rm B\kern-.05em{\sc i\kern-.025em b}\kern-.08em
    T\kern-.1667em\lower.7ex\hbox{E}\kern-.125emX}}
\begin{document}

\title{Vision-Based Agile Landing on Turbulent Waters

\thanks{Manuscript received: May 22, 2026; Revised Jul. 31, 2026; Accepted Aug. 23, 2026. This paper was recommended for publication by Editor Giuseppe Loianno upon evaluation of the Associate Editor and Reviewers' comments. Experiments were performed at the DTU Autonomous Systems Test Arena (DTUASTA20).}

\thanks{\textsuperscript{1}D. Angelis and E. Boukas are with the Department of Electrical and Photonics Engineering, Technical University of Denmark, Kgs. Lyngby, Denmark. (e-mail: diang@dtu.dk)}

\thanks{\textsuperscript{2}L. Bauersfeld and D. Scaramuzza are with the Robotics and Perception Group, Department of Informatics, University of Zurich, Zurich, Switzerland.}

\thanks{Digital Object Identifier (DOI): see top of this page.}
}

\author{
\IEEEauthorblockN{
Dimosthenis Angelis\textsuperscript{1},
Leonard Bauersfeld\textsuperscript{2},
Davide Scaramuzza\textsuperscript{2},
and Evangelos Boukas\textsuperscript{1}
}}

\IEEEaftertitletext{%
\vspace{-2\baselineskip}
\begin{center}
    \includegraphics[width=0.16\textwidth]{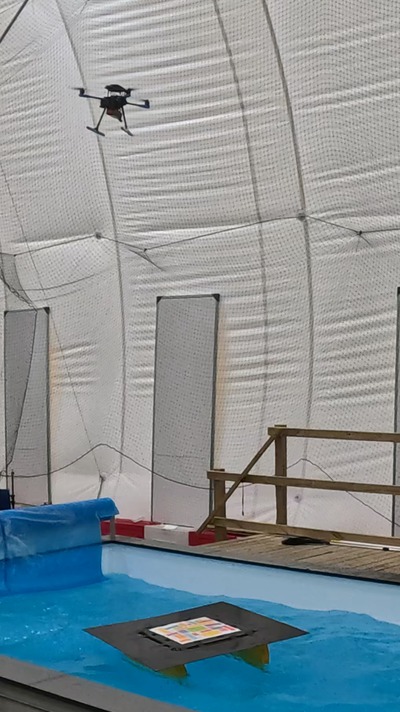}\hfill
    \includegraphics[width=0.16\textwidth]{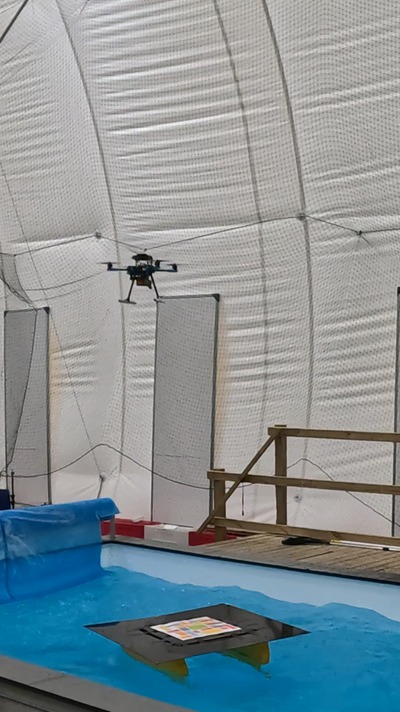}\hfill
    \includegraphics[width=0.16\textwidth]{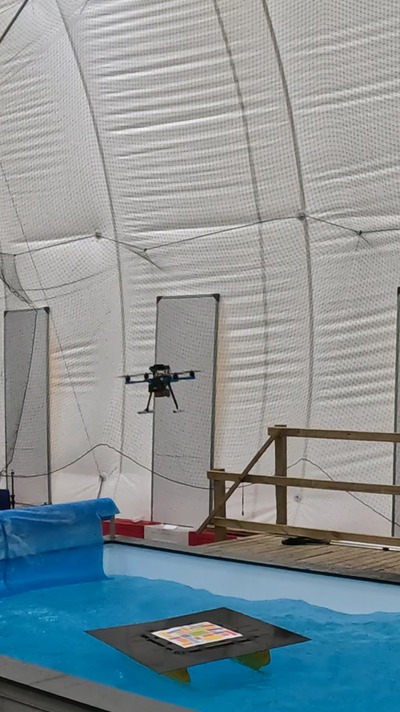}\hfill
    \includegraphics[width=0.16\textwidth]{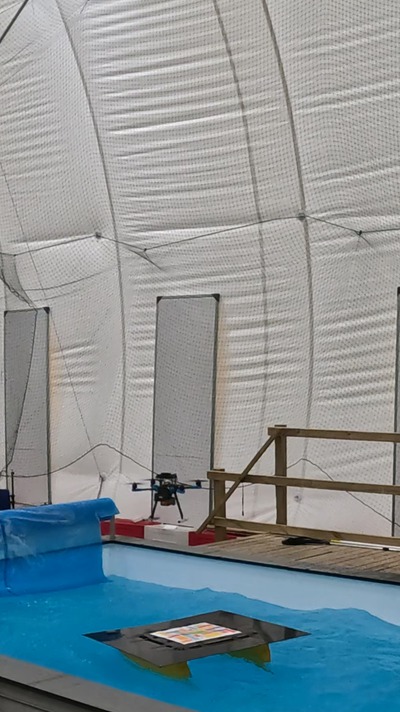}\hfill
    \includegraphics[width=0.16\textwidth]{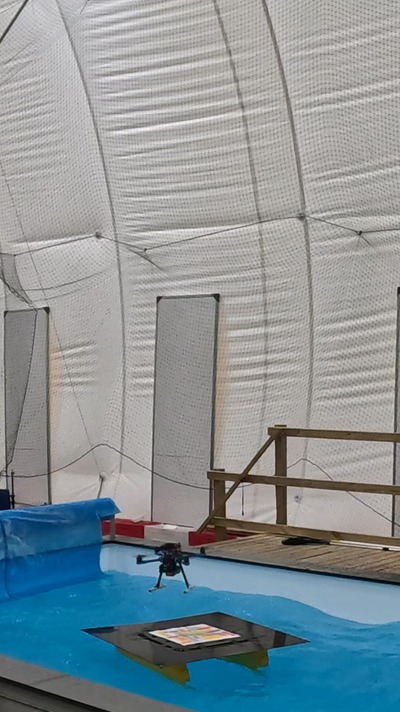}\hfill
    \includegraphics[width=0.16\textwidth]{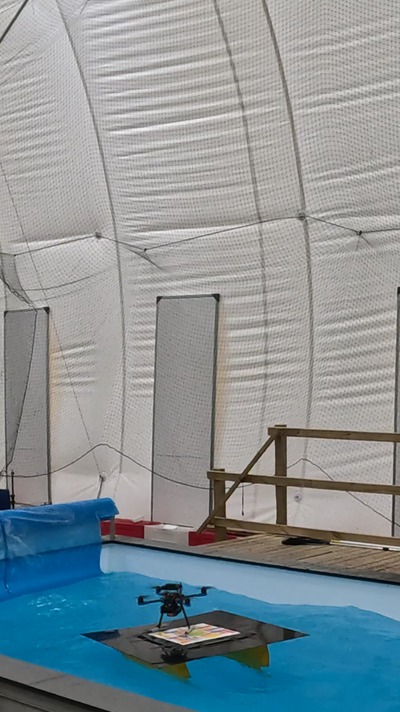}
    \captionof{figure}{A quadrotor lands autonomously on a floating platform in a pool subject to manually generated waves. The landing pad uses a randomly generated target, and all sensing and computation are performed onboard. \\}
    \label{fig:intro}
\end{center}
\vspace{-0.8\baselineskip}
}

\maketitle

\begin{abstract}
  Autonomous landing of Unmanned Aerial Vehicles on maritime vessels is challenging due to the coupled motion of the vehicle and landing platform in open-sea conditions. This paper presents a reinforcement-learning-based approach for autonomous multirotor landing on moving maritime platforms without requiring explicit platform-state observations or estimation during deployment. The proposed method uses multirotor state measurements together with local visual features, consisting of keypoints and associated descriptors extracted from the landing surface, to predict attitude and thrust commands. These commands are tracked by a conventional low-level controller. The policy is trained in simulation using synthetic keypoints with randomly generated normalized descriptors, enabling zero-shot deployment with different local feature extractors onboard the UAV. We evaluate the method in a realistic simulator and show that it outperforms a state-of-the-art Model Predictive Control baseline under platform motions corresponding to “Very Rough” sea conditions. Finally, we perform extensive real-world experiments, demonstrating autonomous onboard landing using two different local feature extractors. To the best of our knowledge, this is the first approach for agile multirotor landing on maritime platforms in turbulent waters that does not rely on an explicit platform-state during deployment.
 \end{abstract}


\vspace{-3mm}
\section{Introduction}
\label{sec:introduction}

\IEEEPARstart{U}{tilization} of Unmanned Aerial Vehicles (UAVs) in maritime operations has increased in recent years due to their effectiveness in search and rescue \cite{doherty_uav_2007, feraru_towards_2020, angelis_uav_2024}, wildlife monitoring \cite{fortuna_using_2013}, and surveillance \cite{peti_aerial_2025}. Their elevated viewpoint and superior agility, compared to surface vehicles, make them particularly effective for detecting objects on the water surface. However, despite their operational advantages, UAVs are constrained by their limited endurance and often require a carrier surface vehicle to transport them to the area of interest for deployment and recovery. This requirement highlights the need for autonomous, agile landing methods that can support efficient and reliable UAV operations from moving maritime platforms. 

The task of landing a multirotor on a 6-Degree-of-Freedom (6-DoF) platform is challenging due to the harsh nature of the sea and the requirements of a successful landing. Traditionally, these missions require the UAV to be able to detect and predict the movement of the landing platform, either directly through cameras that detect predefined geometrical targets on the platform surface, or indirectly by communicating with the landing platform itself. The UAV can then use this information to control its descent and perform the landing. This task becomes increasingly more difficult when the geometric targets are visually degraded due to environmental or operational factors. To the best of our knowledge, no prior approach has demonstrated agile multirotor landing on maritime platforms in turbulent waters without relying on an explicit platform-state representation during deployment.

In this paper, we present a Reinforcement-Learning-based (RL) landing paradigm, where an agent learns how to safely land a multirotor on a 6-DoF landing platform, without requiring explicit platform-state information during deployment. To do so, the agent receives multirotor state information, containing its orientation, linear velocity and angular velocity, and images from its downward-facing camera for two consecutive time steps. The method relies on sparse local features, consisting of keypoints and descriptors extracted from the platform surface in two consecutive images, to implicitly infer the relative motion of the platform with respect to the UAV. It then uses this information to predict the next attitude and thrust commands. In Figure~\ref{fig:intro}, a quadrotor lands autonomously on a floating platform in a pool subject to manually generated waves, using a randomly generated landing target, with all sensing and computation performed onboard. 

The contributions of this paper are as follows: 
\begin{itemize}
  \item proposal of a novel method for landing a multirotor on a moving articulating platform without requiring explicit platform-state information during deployment, under minor, moderate and extreme platform motions,
  \item zero-shot deployment of an RL agent by designing a sparse-feature-based RL policy that can operate with different local feature extractors through a shared normalized feature interface, validated in more than 400 real-world trials, 
  \item outperforming a state-of-the-art Model Predictive Control (MPC) method in a realistic simulator, under platform motions corresponding to conditions up to ``Very Rough'', as described in \cite{prochazka_model_2024}.
\end{itemize}

\section{Related work}
\label{sec:related-work}

Visual-servoing landing methods typically use structured visual targets, such as AprilTags \cite{olson_apriltag_2011}, ArUco markers \cite{garrido-jurado_automatic_2014} or helipad targets, to estimate the relative pose and motion of the multirotor to the landing platform. In the maritime landing setting, Image-Based Visual Servoing (IBVS) has been integrated with vessel velocity estimation \cite{cho_autonomous_2022} and online MPC \cite{yang_robust_2025} to maintain target visibility during the landing maneuver. These methods rely on structured landing targets to perform successful landings, whose detection can degrade in real-world scenarios where the visibility is limited.

Model-based methods try to predict landing platform motions and generate trajectories that satisfy touchdown conditions. \cite{falanga_vision-based_2017} proposes a nonlinear controller paired with a vision-based platform detection and estimation algorithm, to land on a horizontally moving platform. \cite{gupta_landing_2023} uses an Euler-based MPC controller coupled with an online axis decomposition predictor of vessel movements to land a UAV on an Unmanned Surface Vehicle (USV), when the USV tilt angle is less than $5\degree$. This work is extended by \cite{prochazka_model_2024}, where an online MPC-based trajectory generation framework continuously replans trajectories based on the predicted USV poses to land faster and more robustly, utilizing fusion of sensor readings onboard the USV and the UAV. We evaluate our method against MPC-NE which, unlike \cite{prochazka_model_2024}, does not rely on sensor measurements from the USV and is directly comparable to our method.

\begin{figure*}[!t]
  \vspace{-2mm}
\centering
\makebox[\textwidth][c]{%
    \includegraphics[width=1.05\textwidth]{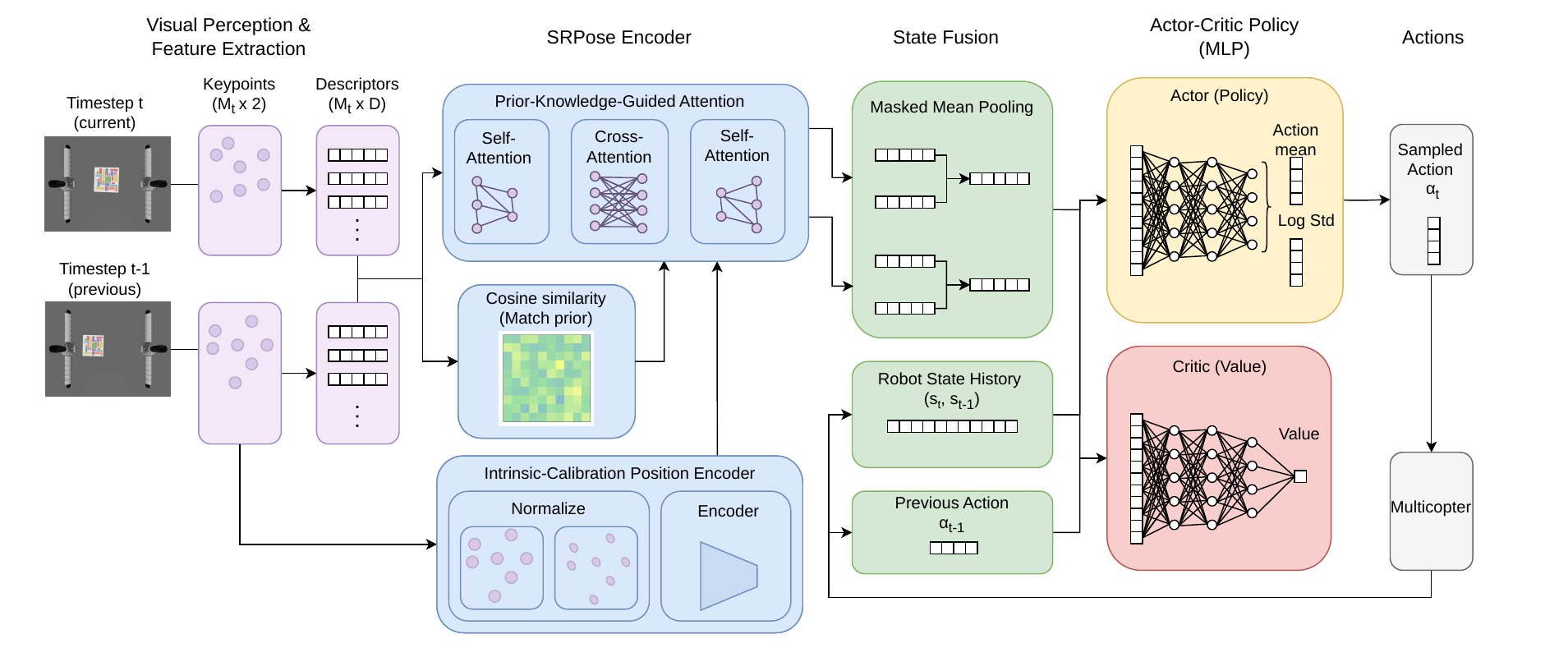}%
}
\caption{Overview of the proposed approach. Sparse local features are extracted from consecutive camera images and encoded together with multirotor state history and the previous action to predict attitude and thrust commands.}
\label{fig:overview}
\vspace{-4mm}
\end{figure*}

RL-based landing methods have also been applied to moving-platform landing with real-world experiments. Existing approaches include Deep Deterministic Policy Gradients algorithms for landing on vertically moving platforms \cite{rodriguez-ramos_deep_2019}, Deep Reinforcement Learning (DRL) policies trained with engineered visual references for landing on 6-DoF motion and wind disturbances \cite{lee_deep_2021}, and hybrid architectures that augment PID controllers with corrective feedback \cite{wu_deep_2022}. These methods use reinforcement learning to land UAVs on moving platforms, but their experiments are limited to zero to small tilt angles. 

\vspace{-3mm}

\section{Methodology}
\label{sec:methodology}

An overview of the proposed method is shown in Figure~\ref{fig:overview}. We train a sensorimotor controller that maps current and previous sparse visual observations, multirotor states, and the previous action to attitude and thrust commands. These commands are used by the low-level attitude controller of our simulated and real multirotors to land on a moving platform. 

\vspace{-3mm}
\subsection{Task Formulation}

We formulate autonomous landing on a moving platform as a Partially Observable Markov Decision Process (POMDP), in which the multirotor must approach the platform, align its motion and attitude with the platform motion, and complete a controlled touchdown. Since the state of the moving platform is not directly available to the policy, the agent must infer the relative motion of the platform from onboard state measurements and temporally adjacent visual observations.

At each control step $t$, the policy outputs a continuous action
\begin{equation}
  a_t = \left[r_t, p_t, y_t, T_t\right],
\end{equation}
where $r_t$, $p_t$, and $y_t$ denote commanded roll, pitch, and yaw, respectively, and $T_t$ is the collective thrust command. These commands are passed to the low-level attitude controller of the multirotor, which tracks the commanded attitude and thrust.

The observation at time step $t$ is defined as: 
\begin{equation}
  o_t = \left[
    M_t, 
    M_{t-1}, 
    s_t,
    s_{t-1}, 
    a_{t-1}
  \right],
\end{equation}
where $M_t$ and $M_{t-1}$ are sparse local features extracted from images $I_t$ and $I_{t-1}$ from the downward-facing camera, $s_t$ and $s_{t-1}$ are multirotor states, and $a_{t-1}$ is the previous action. The state is defined as:
\begin{equation}
  s_{t} = \left[
    \mathbf{v}_t, 
    \mathbf{q}_t,
    \boldsymbol{\omega}_t
  \right], 
\end{equation}
where $\mathbf{v}_t$ is the linear velocity of the multirotor, $\mathbf{q}_t$ is its orientation, and $\boldsymbol{\omega}_t$ is its angular velocity. Camera images and state measurements are collected at a fixed frequency.

The image observations are passed through the vision pipeline to produce a visual feature representation $F_t$. The policy therefore receives the concatenated vector
\begin{equation}
  x_t =
  \left[
  F_t,
  s_t,
  s_{t-1},
  a_{t-1}
  \right],
\end{equation}
and predicts the next action $a_t$.

\subsubsection{Platform Motion Model}

In ship theory, the motion of a ship can be approximated as oscillatory motion with damping effects \cite{ueng_ship_2008}. Based on this observation, we model the landing platform motion as sinusoidal excitation along the heave, roll, pitch, and yaw axes. For each excited DoF $i$, the platform motion is given by
\begin{equation}
  \eta_i(t) =
  A_i \sin \left(2 \pi f t + \phi_i \right),
\label{eq:sinusoidal}
\end{equation}
where $A_i$ is the motion amplitude, $f$ is the oscillation frequency, and $\phi_i$ is the phase offset. The amplitudes are sampled independently for each axis, while the frequency is shared across the excited axes.

\subsubsection{Termination Conditions}

An episode is considered successful when the multirotor reaches the landing region with sufficiently small relative position, orientation, and velocity errors with respect to the platform. The success mask is defined as: 
\begin{equation}
\label{eq:success}
\mathbb{M}_{\mathrm{suc}} =
\begin{cases}
1, &
\begin{aligned}
\text{if } \quad
& d_{xy} < 1\,\mathrm{m} \;\land\; \Delta p_z < 0.2\,\mathrm{m}
\\
& \land\; \Delta\theta_o < 10^\circ \;\land\; \Delta v < 1.5\,\mathrm{m/s},
\end{aligned}
\\
0, & \text{otherwise,}
\end{cases}
\end{equation}
where $d_{xy}$ is the horizontal distance, $\Delta p_z$ is the absolute vertical position error, $\Delta\theta_o$ is the orientation error, and $\Delta v = \lVert \mathbf{v}_r - \mathbf{v}_p \rVert_2$ is the linear velocity error between the multirotor and the platform frames of reference.

An episode is considered failed if the multirotor moves too far away from the platform or it collides with it with excessive force. In cases where the multirotor contacts the platform with low impact force but does not satisfy the success criteria, the episode continues, allowing the multirotor to recover and attempt the landing again. The failure mask is defined as: 
\begin{equation}
\mathbb{M}_{\mathrm{fail}} =
\begin{cases}
1, &
\text{if } d_t > 10\,\mathrm{m}
\;\lor\;
F_{\mathrm{crs}} > 10\,\mathrm{N},
\\
0, & \text{otherwise,}
\end{cases}
\end{equation}
where $d_t$ is the Euclidean distance between the multirotor and the platform center, and $F_{\mathrm{crs}}$ is the measured contact force during collision.

The episode terminates when either $\mathbb{M}_{\mathrm{suc}}=1$, $\mathbb{M}_{\mathrm{fail}}=1$, or when the episode reaches the maximum horizon $T_{\max}$. Episodes that reach $T_{\max}$ without satisfying either terminal condition are counted as timeouts.

\subsubsection{Reward shaping}

The reward function consists of dense shaping terms and terminal terms. The dense terms encourage the multirotor to approach the platform, maintain useful visual observations, reduce relative motion errors, and avoid unsafe configurations. The terminal terms reward successful landings and penalize failed attempts. The reward parameter values used during training are listed in Table~\ref{tab:reward_params}.

The total reward at time step $t$ is
\begin{equation}
\begin{aligned}
\mathcal{R}_t ={}&
r^{\mathrm{app}}_t
+
r^{\mathrm{perc}}_t
+
r^{\mathrm{smooth}}_t
\\
&+
r^{\mathrm{pos}}_t r^{\mathrm{align}}_t
+
r^{\mathrm{term}}_t ,
\end{aligned}
\end{equation}
where $r^{\mathrm{app}}_t$ is the approach reward, $r^{\mathrm{perc}}_t$ is the perception reward, $r^{\mathrm{smooth}}_t$ penalizes abrupt commands, $r^{\mathrm{align}}_t$ rewards velocity and attitude alignment near the platform, $r^{\mathrm{pos}}_t$ acts as a proximity gate, and $r^{\mathrm{term}}_t$ contains the terminal success and failure rewards.

\paragraph{Approach reward}
The approach reward encourages progress toward the platform:
\begin{equation}
  r^{\mathrm{app}}_t
  =
  \lambda_1 \left(d_{t-1} - d_t\right)
  +
  \lambda_2 e^{-\lambda_3 \theta_v^2},
\end{equation}
Here, $\theta_v$ is the angle between the multirotor velocity vector and the direction vector from the multirotor to the platform center. The first term rewards reduction in distance, while the second term encourages the velocity direction to point toward the platform center.

\paragraph{Perception reward}
To keep the landing platform inside the camera field of view, we reward alignment between the camera optical axis and the platform center:
\begin{equation}
  r^{\mathrm{perc}}_t
  =
  \lambda_4 e^{-\lambda_5 \theta_c^2},
\end{equation}
where $\theta_c$ is the angle between the optical axis of the downward-facing camera and the direction vector from the camera to the platform center.

\paragraph{Command smoothness penalty}
To discourage aggressive changes in the commanded attitude and thrust, we penalize action differences between consecutive time steps:
\begin{equation}
  r^{\mathrm{smooth}}_t
  =
  \lambda_6
  \left\lVert a_t - a_{t-1} \right\rVert_2 c_t ,
\end{equation}
where $\lambda_6 < 0$ and $c_t$ is the curriculum progress factor. This penalty becomes more influential as the curriculum progresses.

\paragraph{Near-platform alignment reward}
Once the multirotor approaches the platform, the policy is encouraged to match the platform velocity and orientation. This is achieved using:
\begin{equation}
  r^{\mathrm{align}}_t
  =
  \lambda_{7}e^{-\lambda_{8} \Delta v^2}
  +
  \lambda_{9}e^{-\lambda_{10} \Delta\theta_o^2}
  +
  r^{\mathrm{col}}_t, 
\end{equation}
where
\begin{equation}
  r^{\mathrm{col}}_t
  =
  \lambda_{11}
  \min
  \left(
  \left(
  \left|\Delta p_z\right| + \Delta p_z
  \right)
  \cos\Delta\theta_o
  -
  \lambda_{12},
  0
  \right).
\end{equation}
The collision-avoidance term $r^{\mathrm{col}}_t$ penalizes configurations in which the multirotor is close to the platform while being poorly aligned, which could result in propeller-platform contact.

The alignment reward is multiplied by a proximity term:
\begin{equation}
  r^{\mathrm{pos}}_t
  =
  e^{-\lambda_{13} d_t^2}.
\end{equation}
This ensures that precise velocity and orientation matching are emphasized primarily when the multirotor is close to the landing platform.

\paragraph{Terminal reward}
The terminal reward consists of a success bonus and a failure penalty:
\begin{equation}
  r^{\mathrm{term}}_t
  =
  r^{\mathrm{suc}}_t
  +
  r^{\mathrm{fail}}_t ,
\end{equation}
where
\begin{equation}
  r^{\mathrm{suc}}_t
  =
  \mathbb{M}_{\mathrm{suc}}
  \left(
  \lambda_{14}\frac{T_{\max}-t}{T_{\max}}
  +
  \lambda_{15}(T_{\max}-t)
  \right),
\end{equation}
\begin{equation}
  r^{\mathrm{fail}}_t
  =
  \lambda_{16}\mathbb{M}_{\mathrm{fail}}.
\end{equation}
The success reward is scaled by the remaining episode time, encouraging the agent to complete the landing efficiently rather than hovering near the platform. The failure term penalizes crashes and out-of-bounds states.

\begin{table}[t]
\centering
\caption{Reward-function parameter values.}
\label{tab:reward_params}
\scriptsize
\setlength{\tabcolsep}{4pt}
\renewcommand{\arraystretch}{0.88}
\begin{tabular}{@{}cc|cc|cc|cc@{}}
\toprule
Param. & Value & Param. & Value & Param. & Value & Param. & Value \\
\midrule
$\lambda_1$ & 1e0   & $\lambda_5$ & 1e1   & $\lambda_9$    & 1e-1 & $\lambda_{13}$ & 5e-1 \\
$\lambda_2$ & 5e-2  & $\lambda_6$ & -1e-1 & $\lambda_{10}$ & 1e1  & $\lambda_{14}$ & 1e1  \\
$\lambda_3$ & 5e0   & $\lambda_7$ & 2e-2  & $\lambda_{11}$ & 1e-1 & $\lambda_{15}$ & 2e-2 \\
$\lambda_4$ & 2e-2  & $\lambda_8$ & 5e0   & $\lambda_{12}$ & 2e-1 & $\lambda_{16}$ & -1e0  \\
\bottomrule
\end{tabular}
\end{table}

\vspace{-2mm}
\subsection{Vision pipeline}

To improve robustness to visual domain shift, we provide the policy with a sparse visual-feature representation rather than raw RGB images. Instead of using segmentation masks as input to our network as prior works do \cite{bartolomei_semantic-aware_2021, kaufmann_champion-level_2023, geles_demonstrating_2024}, we use sparse local features. We use the term sparse local features to refer to keypoint coordinates together with their associated descriptors. Segmentation-based representations can be sensitive to boundary quality, illumination, and appearance changes between simulation and real maritime scenes. Sparse local features provide a lower-dimensional abstraction that emphasizes repeatable local structure rather than dense appearance.

SRPose \cite{yin_srpose_2025} provides a sparse keypoint-based encoder for two-view relative pose estimation. We adopt the SRPose encoder as the visual abstraction module for our policy, with the modification that all descriptors are L2-normalized before being processed by the network. At each time step $t$, sparse keypoints and descriptors are extracted from the current and previous images, $I_t$ and $I_{t-1}$. The keypoint coordinates are normalized using the camera intrinsics, while the descriptors are L2-normalized so that the cosine-similarity depends on descriptor direction rather than descriptor magnitude.

The normalized descriptors are used to construct a match prior between the two views, and the normalized keypoint coordinates are passed through the position encoder to obtain positional embeddings. The resulting descriptor and positional representations are processed by the SRPose attention encoder, which combines self-attention within each image with prior-knowledge-guided cross-attention across the two images. This produces a compact feature representation that captures implicit correspondences between temporally adjacent observations. The resulting visual feature vector $F_t$ is then provided to the policy network, allowing the policy to infer relative motion cues between the multirotor and the landing platform without requiring platform-state estimation or prediction.

L2 normalization provides a common descriptor interface across training and deployment and facilitates transfer between local feature extractors. During deployment, the local feature extractor must produce descriptor vectors with the same dimensionality as those used during training, which we normalize before passing to the SRPose encoder. In this formulation, the descriptors are used to define correspondence priors between keypoints across images, rather than to provide visual or semantic appearance information to the control policy. This decouples the control policy from raw image appearance and reduces the vision-induced sim-to-real gap, while still allowing different local feature extractors to be used within the same sparse-feature interface.

\vspace{-4mm}
\subsection{Policy}

The policy is implemented as a Gaussian actor-critic architecture with separate MLPs for the actor and critic. On each time step $t$, both networks receive the concatenated input $x_t$. The actor outputs the mean of a Gaussian action distribution, while the critic outputs a scalar value estimate. Both networks are two-layer MLPs with 512 hidden units per layer and $\tanh$ activations. No privileged information is provided to the critic.

\vspace{-3mm}

\subsection{Training details}
\label{seq:training-details}

We conduct training experiments in Aerial Gym \cite{kulkarni2025aerial}, a highly parallelized GPU-based simulator for aerial vehicles. This enables us to run multiple episodes simultaneously, reducing the time needed for training. The SRPose encoder, policy network, and value network are trained jointly using PPO \cite{schulman_proximal_2017}, with the output of the image-based local feature extraction module replaced by synthetic projected platform features. The PPO hyperparameters are listed in Table~\ref{tab:training_params}.

\begin{table}[t]
\centering
\caption{PPO training hyperparameters.}
\label{tab:training_params}
\scriptsize
\setlength{\tabcolsep}{4pt}
\renewcommand{\arraystretch}{0.88}
\begin{tabular}{@{}l c l c@{}}
\toprule
Parameter & Value & Parameter & Value \\
\midrule
Total time steps & $10^8$ & Rollout steps & $256$ \\
Mini-batches & $16$ & Update epochs & $1$ \\
Learning rate & $3 \times 10^{-4}$ & Optimizer & Adam \\
Discount $\gamma$ & $0.98$ & GAE $\lambda$ & $0.95$ \\
Clip coefficient & $0.2$ & Entropy coeff. & $10^{-3}$ \\
Value-loss coeff. & $2.0$ & Max grad. norm & $1.0$ \\
\bottomrule
\end{tabular}
\end{table}

To support transfer to the real multirotor, we model the vehicle in simulation and tune its low-level attitude and thrust response to match the real multirotor response to commanded actions. At the beginning of each episode, the multirotor is initialized above the landing platform such that the platform is fully or partially visible from the downward-facing camera. The camera pose is randomized within small perturbation bounds to improve robustness to extrinsic calibration errors.

At the beginning of each episode, the platform is initialized at a fixed nominal position. The amplitudes of the heave, roll, pitch, and yaw motions are sampled from uniform distributions, with roll and pitch sampled from $[-\pi/6,\pi/6]$ rad, yaw from $[-\pi/16,\pi/16]$ rad, and heave from $[0.0,5.0]$ m. A single oscillation frequency is sampled from $[0,0.5]$ Hz at the beginning of each episode and shared across all excited degrees of freedom. These ranges are chosen to expose the policy to a wide range of short-horizon platform motions and accelerations, and expand the learned space.

We use curriculum learning to improve training stability and sample efficiency. Training begins with a static landing platform. Once the success rate exceeds $90\%$, roll, pitch, and yaw amplitudes and frequencies are sampled from the predefined ranges, while the maximum heave amplitude is kept at $0$ m. Subsequently, every $4096$ episodes, the maximum heave amplitude is increased by $0.5$ m if success rate exceeds $60\%$ and is decreased by $0.25$ m if success rate falls below $20\%$.

For each episode, $25$ points are sampled on the platform surface using a uniform distribution. Each point is assigned a fixed 64-dimensional descriptor vector sampled from a normal distribution. At each time step, the visible points are projected into the camera frame and used as sparse local-feature observations. To emulate imperfect feature extraction, $20\%$ of the descriptors are randomly dropped at each time step, and noise is added to the remaining descriptors. This randomization encourages robustness to missing features, descriptor noise, and changes in the deployment local feature extractor.

\begin{figure}[!t]
\centering
\includegraphics[width=\columnwidth]{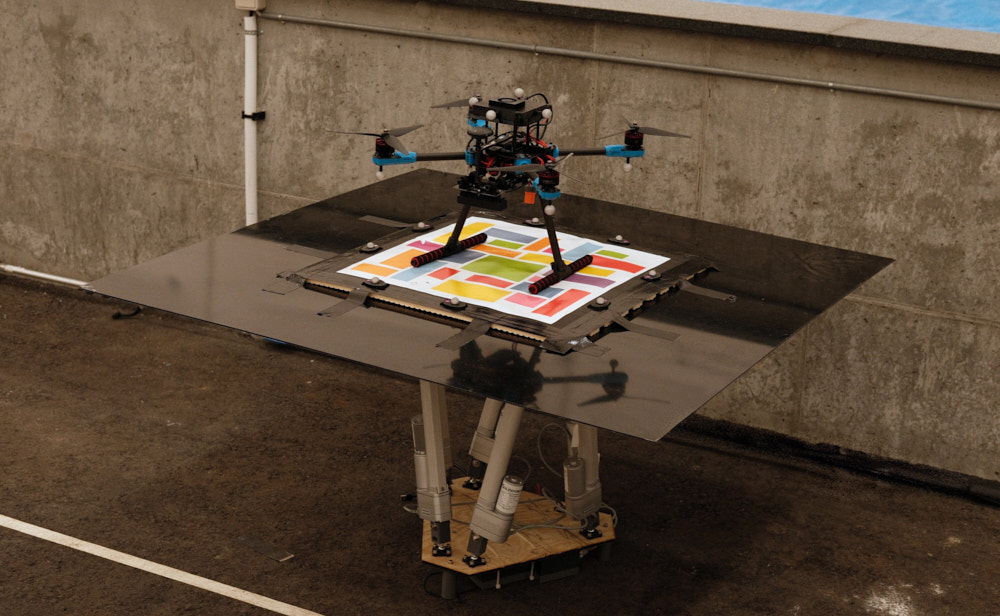}
\caption{Illustration of quadrotor and 6-DoF Stewart Platform used for experiments. The pattern on top of the landing platform is randomly generated and not seen during training. The markers on the quadrotor and platform are used for offline logging and are not part of the deployed pipeline.}
\label{fig:real-world-setup}
\end{figure}

\vspace{-2mm}
\section{Experiments}
\label{sec:experiments}

We evaluate the proposed method in both a ROS-based Gazebo simulation environment and real-world experiments. The deployment pipeline consists of three components: landing-platform detection, sparse local feature extraction, and the trained landing policy.

\vspace{-3mm}
\subsection{Experiments in Simulations}

\begin{figure}[!t]
\centering
\includegraphics[width=\columnwidth]{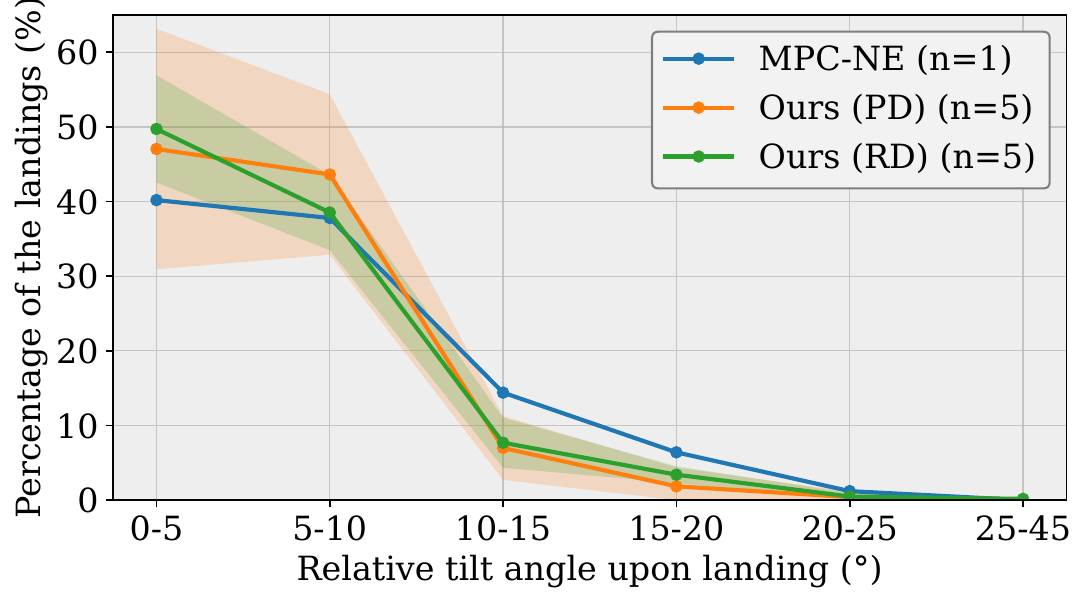}
\caption{Distribution of relative tilt angle upon landing (random platform motion). Solid lines show the mean across seeds. Shaded regions are 95\% confidence intervals over the $n$ seeds indicated in the legend. MPC-NE has a single run, so no interval is shown.}
\label{fig:simulation-results}
\end{figure}

We first evaluate the proposed method in a realistic ROS-based Gazebo simulation environment. We compare against MPC-NE \cite{gupta_landing_2023}, a state-of-the-art, non-cooperative, moving-platform landing method for multirotors. MPC-NE uses fiducial markers placed on the landing platform to detect and predict the platform motion. To separate the effect of the visual target representation from the landing policy, we evaluate two variants of the proposed method: one using a predefined AprilTag-based feature layout and one using a randomly generated visual pattern on the landing platform.

In the predefined-descriptor setting, denoted as Ours (PD), the policy is trained using $25$ uniformly distributed points arranged in a symmetric $5 \times 5$ pattern. Each point is associated with a predefined randomly generated descriptor. During deployment, an AprilTag detector is used to localize the tag on the landing platform, and the same $5 \times 5$ keypoint layout and descriptor assignment used during training are applied to the detected tag region. The resulting keypoint-descriptor pairs are then passed to the policy for inference.

In the random-descriptor setting, denoted as Ours (RD), the policy is trained as described in Section~\ref{sec:methodology}, using randomly generated keypoints and descriptors. During testing, the landing-platform region is detected, and A-KAZE \cite{alcantarilla_fast_2013} is used to extract keypoint-descriptor pairs from the randomly generated visual pattern on the platform. The $25$ keypoints with the highest detector response are selected and passed to the policy.

For each descriptor setting, we train five models using different random seeds. All methods are evaluated independently over $500$ landing trials. Platform motion is generated along the roll, pitch, and heave axes by superimposing five sinusoidal components defined in Equation~\ref{eq:sinusoidal}. The oscillation amplitudes are sampled to limit the maximum tilt about each rotational axis to $30\degree$ and the vertical displacement to $5$ m. The vertical displacement limit is selected to reflect the scale of the $4$--$6$ m wave-height range associated with ``Very Rough'' sea conditions in \cite{prochazka_model_2024}. Figure~\ref{fig:simulation-results} shows the distribution of the relative touchdown angle between the multirotor and the platform. For our methods, confidence intervals computed across the different random seeds are also reported. On average, our methods produce better tilt alignment with the landing platform during touchdown. 

While the touchdown-angle distribution provides insight into attitude alignment at contact, it does not fully characterize landing quality because it does not account for the relative touchdown velocity or relative position. Therefore, we additionally report Success rate, Crash rate, and Maneuver Time (MT) in Table~\ref{tab:simulation_results}. Success is evaluated using the criteria defined in Equation~\ref{eq:success}. For MPC-NE, we exclude the estimator convergence time and measure only the landing maneuver time. 

Table~\ref{tab:simulation_results} shows that Ours (PD) achieves the highest success rate and the lowest crash rate. Compared with MPC-NE, Ours (PD) improves the success rate by $25.61$ percentage points on average and approximately halves the average maneuver time. Ours (RD) achieves the shortest average maneuver time by sacrificing $11.13$ percentage points in average success rate, compared to Ours (PD). In the predefined-descriptor setting, the policies tend to take more corrective actions close to the platform, achieving the best success rates. Overall, the results indicate that the proposed policies enable faster and more successful landings than MPC-NE under the evaluated simulation conditions.

\begin{table}[!t]
\centering
\caption{Simulation results for different landing methods.}
\label{tab:simulation_results}
\setlength{\tabcolsep}{4pt}
\renewcommand{\arraystretch}{0.88}
\resizebox{\columnwidth}{!}{%
\begin{tabular}{@{}l c c c@{}}
\toprule
\textbf{Method} & \textbf{Success (\%)} & \textbf{Crash (\%)} &  \textbf{Maneuver Time (s)} \\
\midrule
MPC-NE & 48.60 & 51.40 & 9.06 $\pm$ 9.14$^{\dagger}$ \\
Ours (PD) & \textbf{74.21 $\pm$ 3.93} & \textbf{25.79 $\pm$ 3.93} & 5.02 $\pm$ 2.28 \\
Ours (RD) & 63.08 $\pm$ 11.26 & 36.92 $\pm$ 11.26  & \textbf{4.28 $\pm$ 1.62} \\
\bottomrule
\end{tabular}%
}
\par\smallskip{Values for Ours (PD) and Ours (RD) are mean ± standard deviation across five independently trained seeds. \raggedright\scriptsize $^{\dagger}$ Single-configuration method: spread is over the individual landings of that run rather than across seeds.\par}
\end{table}

\vspace{-3mm}
\subsection{Deployment in the Real World}

\begin{table*}[!t]
\centering
\caption{Real-world experiment results.}
\label{tab:real_world_experiment_results}
\scriptsize
\setlength{\tabcolsep}{3pt}
\renewcommand{\arraystretch}{0.88}
\begin{tabular}{@{}l c c c c c c@{}}
\toprule
\textbf{Experiment type} & \textbf{\# Experiments} & \textbf{Tilt angle violation (\%)} & \textbf{Velocity violation (\%)} & \textbf{Touchdown Success (\%)} & \textbf{General Success (\%)} & \textbf{MT (s)} \\
\midrule
Stewart platform (SURF) & 181 & 41.44 & 4.97 & 57.46 & 96.13 & 4.95 $\pm$ 7.71\\
Stewart platform (AKAZE)  & 177 & 39.55 & 3.39 & 59.32 & 97.74 & 3.50 $\pm$ 1.08\\
Pool (AKAZE) & 53 & 3.77 & 7.54 & 90.57 & 98.11 & 2.90 $\pm$ 1.56\\
\bottomrule
\end{tabular}
\end{table*}

\begin{table*}[!t]
\centering
\caption{MLP-probe decodability of the platform's pose and velocity in the robot frame from the frozen visual feature (\textit{$F_t$}), from a single frame with the previous frame ablated
(\textit{$F_t$, 1\,img}), and from the full actor input (\textit{$x_t$}).}
\label{tab:probe_mlp}
\scriptsize
\setlength{\tabcolsep}{5pt}
\renewcommand{\arraystretch}{0.88}
\begin{tabular}{@{}l cc cc cc cc@{}}
\toprule
 & \multicolumn{2}{c}{\textbf{Position}} & \multicolumn{2}{c}{\textbf{Orientation}}
 & \multicolumn{2}{c}{\textbf{Lin. velocity}} & \multicolumn{2}{c}{\textbf{Ang. velocity}} \\
\cmidrule(lr){2-3}\cmidrule(lr){4-5}\cmidrule(lr){6-7}\cmidrule(lr){8-9}
\textbf{Encoder input} & $\|$RMSE$\|$\,[m] & $R^2$ (x/y/z) & mean/med\,[$^\circ$] & $R^2$ (r/p/y)
 & $\|$RMSE$\|$\,[m/s] & $R^2$ (x/y/z) & $\|$RMSE$\|$\,[rad/s] & $R^2$ (r/p/y) \\
\midrule
\multicolumn{9}{@{}l}{\emph{PD}}\\
Trained ($F_t$, 1\,img)  & 0.179 & 1.00/1.00/1.00 & 2.53/1.29 & 0.95/0.95/0.94 & 0.639 & 0.43/0.68/0.56 & 0.485 & 0.15/0.20/0.29 \\
Trained ($F_t$)          & 0.281 & 1.00/1.00/0.99 & 2.25/0.99 & 0.95/0.96/0.93 & 0.408 & 0.76/0.92/0.77 & 0.224 & 0.84/0.79/0.79 \\
Trained ($x_t$)          & 0.248 & 1.00/1.00/0.99 & 2.18/1.02 & 0.95/0.96/0.95 & 0.156 & 1.00/1.00/0.95 & 0.122 & 0.96/0.91/0.94 \\
\addlinespace[2pt]
\multicolumn{9}{@{}l}{\emph{RD}}\\
Trained ($F_t$, 1\,img)  & 0.506 & 0.90/0.88/0.93 & 11.20/8.88 & 0.13/0.15/0.05 & 0.483 & 0.43/0.45/0.64 & 0.626 & 0.01/0.00/-0.02 \\
Trained ($F_t$)          & 0.371 & 0.94/0.93/0.97 & 9.27/7.55  & 0.44/0.42/0.20 & 0.298 & 0.78/0.77/0.87 & 0.406 & 0.56/0.64/0.34 \\
Trained ($x_t$)          & 0.374 & 0.94/0.93/0.97 & 7.29/5.82  & 0.59/0.58/0.79 & 0.122 & 0.99/0.99/0.97 & 0.211 & 0.87/0.90/0.92 \\
\bottomrule
\end{tabular}
\end{table*}

We further validate the proposed method through real-world landing experiments. The complete deployment pipeline runs onboard a custom quadrotor equipped with a Jetson Orin Nano. An object detection module first localizes the landing-platform region in the downward-facing camera image. Sparse local features are then extracted inside the detected region using either A-KAZE or SURF \cite{bay_speeded-up_2008}. For both methods, we use non-extended 64-dimensional descriptors. The $25$ features with the highest detector response are selected and passed to the trained policy, which predicts attitude and thrust commands. These commands are tracked by the tuned PX4 low-level controller. The state estimation is performed onboard through PX4 EKF2, using the IMU and GNSS sensors onboard the UAV. The policy communicates with PX4 through ROS~2. Object detection requires $19$ ms, feature extraction requires $25$ ms with A-KAZE or $11$ ms with SURF, and fused encoder-policy inference requires $9$ ms. The resulting end-to-end latency is $53$ ms with A-KAZE and $39$ ms with SURF. The pipeline stages are parallelized, allowing $30$ Hz throughput. 

We first evaluate the method using a 6-DoF Stewart platform~\cite{stewart_platform_1965} with a randomly generated visual pattern attached to the landing pad. The experimental setup is shown in Figure~\ref{fig:real-world-setup}. We perform $358$ landing trials with the platform oscillating in roll and heave at $0.2$ Hz. The maximum roll and heave amplitudes are $20^\circ$ and $0.25$ m, respectively. During deployment, we switch between A-KAZE and SURF while using the same trained policy. This evaluates whether the policy can operate with multiple local feature extractors through the shared sparse-feature interface, without policy retraining. Table~\ref{tab:real_world_experiment_results} reports the real-world deployment results using this setup. The evaluation criteria are the same as those used in simulation. Most failed trials violate the touchdown tilt-angle criterion rather than the velocity criterion. In these cases, the vehicle often remains on the platform after contact instead of rebounding or crashing. We report this outcome as General Success (\%).

\begin{table}[!t]
\centering
\caption{Pool experiment landing-platform 6-DOF motion parameters, showing the significant amplitude $SA$, Peak-to-peak $P2P$, spectral zero-crossing period $\bar{T}$, and spectral bandwidth $\nu$.}
\label{tab:pool-platform-motion}
\scriptsize
\setlength{\tabcolsep}{4pt}
\renewcommand{\arraystretch}{0.88}
\begin{tabular}{@{}l c c c c@{}}
\toprule
\textbf{Motion} & \textbf{SA} & \textbf{P2P} & \textbf{$\bar{T}$ (s)} & \textbf{$\nu$} \\
\midrule
Surge (x) & 2.9\,cm      & 11\,cm      & 1.4 & 0.64 \\
Sway (y) & 19\,cm       & 48\,cm      & 2.0 & 0.18 \\
Heave (z) & 4.4\,cm      & 13\,cm      & 1.2 & 0.48 \\
\midrule
Roll (r) & 8.5$^\circ$  & 25$^\circ$  & 1.1 & 0.36 \\
Pitch (p) & 6.3$^\circ$  & 23$^\circ$  & 1.0 & 0.33 \\
Yaw (y) & 4.3$^\circ$  & 18$^\circ$  & 2.2 & 0.93 \\
\bottomrule
\end{tabular}
\end{table}

Finally, we evaluate the method using the same landing pad mounted on a floating platform in a pool, where manually generated waves induce platform motion, as shown in Figure~\ref{fig:intro}. In these experiments, the same onboard perception-control pipeline used in the Stewart-platform experiments was deployed without retraining. The purpose of this evaluation is to assess the performance of our method under wave-induced floating-platform motion, which produces irregular platform dynamics that differ from the sinusoidal motion model used during training. We perform $53$ landing trials with the platform passively oscillating. The measured motion characteristics are shown in Table~\ref{tab:pool-platform-motion}. The results reported in Table~\ref{tab:real_world_experiment_results} show that the policy is able to generalize to landing platform movements that are not present in training and achieve a touchdown success rate of 90.57\% and general success rate of 98.11\%. 

\vspace{-3mm}
\subsection{Analysis of Learned State Representations}
\label{sec:encoder-ablation}

We evaluate whether the trained SRPose encoder represents information about the landing platform pose and velocity in the robot frame. To do so, we train a simple three-layer MLP probe with hidden layer widths of $64$ and $32$ and Tanh activations, to predict the landing platform position, orientation, linear velocity and angular velocity. The dataset contains 1.5 million simulation steps, excluding samples with less than four visible features, and is split into training and validation sets using an 80/20 ratio. The policy and encoder remain frozen throughout training. We evaluate three inputs for both PD and RD variants: the visual features from one image with the previous frame ablated, the complete two-frame image input, and the full actor input.

The results for both PD and RD methods are shown in Table~\ref{tab:probe_mlp}. In the PD case, position and orientation are accurately decoded from a single image, but velocity is only weakly represented. Using both images substantially improves velocity predictions, indicating that temporal information is required to encode landing platform motion. The full actor input achieves the best performance. In the RD case, we see the same general trend, except that orientation is decoded more accurately only when the full actor input is used. This weaker orientation representation is consistent with the policy's tendency to violate the tilt-angle criterion in the real-world experiments (Table~\ref{tab:real_world_experiment_results}). 

\section{Conclusion}
\label{sec:conclusion}

This paper presented a reinforcement-learning-based approach for autonomous multirotor landing on moving maritime platforms. The proposed method uses sparse local features together with multirotor state information to directly predict attitude and thrust commands that can be tracked by a conventional attitude controller. The simulation experiments demonstrated that the proposed method outperforms the MPC-NE baseline in both success rate and average maneuver time, for sea conditions up to ``Very Rough'', based on the classification in \cite{prochazka_model_2024}. Furthermore, the real-world experiments validated that the same policy can be used zero-shot with different local feature extractors and can run onboard a UAV without offboard sensing, computation, or explicit platform-state information during deployment.

\bibliographystyle{IEEEtran}
\bibliography{references}

\vspace{12pt}

\end{document}